\documentclass[11pt]{article}
\usepackage{coling2020}
\usepackage{times}
\usepackage{url}
\usepackage{latexsym}
\usepackage{graphicx}
\usepackage{comment}
\usepackage{booktabs}
\usepackage{amsmath}
\usepackage{multirow}

\newtheorem{proposition}{Proposition}
\newtheorem{definition}{Definition}

\colingfinalcopy 

\title{Meta Sequence Learning for Generating Adequate Question-Answer Pairs}

\author{Cheng Zhang \\
  Department of Computer Science \\
  University of Massachusetts \\
  Lowell 01854, MA \\
  {\tt cheng\_zhang@student.uml.edu} \\\And
  Jie Wang \\
  Department of Computer Science \\
  University of Massachusetts \\
  Lowell 01854, MA \\
  {\tt wang@cs.uml.edu} \\}

\date{}

\begin{document}
\maketitle

\begin{abstract}
Creating multiple-choice questions to assess reading comprehension of a given article
involves generating question-answer pairs (QAPs) on the main points of the document.
We present a learning scheme to generate adequate QAPs 
via meta-sequence representations of sentences.  
A meta sequence is a sequence of vectors comprising semantic and syntactic tags.
In particular,
we devise a scheme called MetaQA to
learn meta sequences from training data to form 
pairs of a meta sequence for a declarative sentence (MD) 
and a corresponding  interrogative sentences (MIs). 
On a given declarative sentence, a trained MetaQA model 
converts it to a meta sequence, 
finds a matched MD,  and  
uses the corresponding MIs and the input sentence to generate QAPs.
We implement MetaQA for the English language using 
semantic-role labeling, 
part-of-speech tagging, and  named-entity recognition,
and show that, trained on a small dataset, 
MetaQA generates efficiently over the official SAT practice reading tests a large number of syntactically and semantically correct QAPs with over 97\% accuracy.
\end{abstract}

\section{Introduction}

In an effort to build an online learning tool for helping students improve reading comprehension, it calls for a system to automatically generate adequate 
multiple-choice questions (MCQs) to assess student's understanding of a given article's main points. 
An article's main points include
\textsl{direct} and \textsl{derived} points. 
A direct point is expressed in a declarative sentence. A derived point 
is inferred from multiple direct points, 
which could be a causal relation between them, an aggregation over them, or a conclusion drawn from them.

We study automatic generation of question-answer pairs (QAPs) with
an emphasis on the grammatical correctness of the questions 
and the suitability of the answers. By grammatical correctness we mean
that the questions being generated are syntactically and semantically correct and 
conform to what a native speaker would say.
We refer to such QAPs as \textsl{adequate} QAPs.
Other tasks of generating MCQs not addressed in this paper are how to provide adequate
distractors for an answer.

Existing methods on QAP generation are based on handcrafted features or neural networks.
While they have met with certain success
from different perspectives,
the grand challenge of generating adequate QAPs 
still remains. 

We present a new approach to tackling this challenge. 
In particular,
we use a sequence of vectors to represent a sentence, where each vector consists of 
a semantic-role (SR) tag, a part-of-speech (POS) tag, and
other syntactic and semantic tags,
 and we refer to such a sequence as a \textsl{meta sequence}.
We then present a scheme called MetaQA to learn meta sequences 
of declarative sentences and the corresponding interrogative sentences
from a training dataset. 
Combining and removing redundant meta sequences yields a set called MSDIP 
(Meta-Sequence-Declarative-Interrogative Pairs), with each element being
a pair of an MD and corresponding MI(s),
where MD and MI stand for, respectively, a meta sequence for a declarative sentence and
for an interrogative sentence.
A trained MetaQA model generates QAPs for a given declarative sentence $s$ as follows:
Generate a meta sequence for $s$, find a best-matched MD from MSDIP,
generates meta sequences for interrogative sentences according to the corresponding MIs
and the meta sequence of $s$, identifies the meta-sequence answer to each MI,
and coverts them back to text to form a QAP.

We implement MetaQA for the English language using  SR, POS, and
NE (named-entity) tags.
We then train MetaQA using a moderate initial dataset and
show that MetaQA generates efficiently a large number of 
adequate QAPs with an accuracy of 97\% on the official SAT practice reading tests.
These tests contain a large number of declarative sentences in different patterns, and 
there is no match in the initial MSDIP for some of these sentences. After learning interrogative for some of these sentences, MetaQA successfully generate many more adequate QAPs.

The rest of the paper is organized as follows: We describe in Section \ref{sec:2} related work, in Section \ref{sec:3} the details of meta sequence learning. We then present in Section \ref{sec:4} the answer generation. We report evaluation results in Section \ref{sec:5}. Finally, we conclude the paper in Section \ref{sec:6}. 

\section{Related Work} \label{sec:2}

Automatic question generation (QG), first studied by Wolfe \cite{wolfe1976automatic} as a means to aid independent study, has since attracted increasing attentions
in two lines of methodologies: transformative and generative.
 
\paragraph{Transformative methods.}
Transformative methods transform key phrases from a given single declarative sentence into  factual questions.
Existing methods are rule-based on syntax, semantics, or templates.

Syntactic-based methods
follow the same basic strategy: Parse sentences using a syntactic parser to identify key phrases and transform a sentence to a question based on syntactic rules. 
These include methods
to identify key phrases from input sentences and use syntactic rules for 
different types of questions  \cite{varga2010wlv}, 
generate questions and answers using a syntactic parser, a POS tagger, and an NE analyzer \cite{ali2010automation},
transform a sentence into a set of questions using a series of domain-independent rules \cite{danon2017syntactic}, and
generate questions using relative pronouns and adverbs from complex English sentences \cite{khullar2018automatic}. 

Semantic-based methods create questions using
predicate-argument structures and semantic roles \cite{mannem2010question},
semantic pattern recognition  \cite{mazidi2014linguistic}, 
subtopics based on Latent Dirichlet Allocation \cite{chali2015towards}, or
semantic-role labeling 
\cite{flor2018semantic}.

These methods are similar. The only difference is that semantic-based methods use semantic parsing while syntactic-based methods ues syntactic parsing to determine which specific words or phrases should be asked. In a language with many syntactic and
semantic exceptions, such as English, these methods would require substantial manual labor to construct rules.

Template-based methods are for special-purpose applications with built-in templates. Research in this line
devises a Natural Language Generation Markup Language (NLGML) \cite{cai2006nlgml};
uses a phrase structure parser to parse text and construct questions using enhanced XML
 \cite{rus2007experiments};
devise a self-questioning strategy to help children generate questions from narrative fiction \cite{mostow2009generating};
use informational text to enhance 
the self-questioning strategy \cite{chen2009aist};
apply pattern matching, variables, and templates to transform source sentences into questions similar to NLGML \cite{wyse2009generating};
defines a question template as pre-defined text with placeholder variables to be replaced with content from the source text \cite{lindberg2013automatic}; or
incorporates semantic-based methods into a template-based method to support online learning \cite{lindberg2013generating}.

\paragraph{Generative methods.}

Recent advancements of neural-network methodologies 
have shed new light on
generative methods.
For example,
the attention mechanism  \cite{luong-etal-2015-effective} is 
used to determine what content in a sentence should be asked,
and the sequence-to-sequence  \cite{bahdanau2014neural,cho-etal-2014-learning} and the long short-term memory  \cite{Sak2014LongSM}  mechanisms are used to generate each word in a question (see, e.g., \cite{du-etal-2017-learning,duan-etal-2017-question,Harrison_2018,sachan-xing-2018-self}).
These models, however, only deal with question generations
without generating correct answers. 
Moreover, training these models require a dataset comprising over 100K questions.

To address the problem of generating questions without answers, researchers have explored ways to encode a passage (a sentence or multiple sentences) and an answer word (or a phrase) as input, and determine what questions are to be generated for a given answer
\cite{10.1007/978-3-319-73618-1_56,zhao-etal-2018-paragraph,song-etal-2018-leveraging}.
Kim et al. \cite{Kim_2019} pointed out that these models could generate a number of
answer-revealing questions (namely, questions contain in them the corresponding answers). They then
devised a new method
by encoding answers separately, at the expsense of having substantially more parameters. Their experiments show that the BLEU-4 \cite{10.3115/1073083.1073135}, 
METEOR \cite{banerjee-lavie-2005-meteor},
and ROUGE-L \cite{lin-2004-rouge} scores on the questions generated are, respectively,
16.2, 19.92, 43.96, which are
3 to 4 points higher than the earlier results on the same dataset 
\cite{du-etal-2017-learning}.
On top of low accuracy,
it is also unknown whether the questions generated are grammatically correct 
because these measures do not measure grammatical correctness.

\section{Meta Sequence Learning} \label{sec:3}

Our objective is to generate adequate QAPs on a given declarative sentence written in a given language $L$. We 
assume that $L$ has an oracle $O_L$ to
provide syntactic and semantic information on a given sentence.

\begin{enumerate}
\vspace*{-5pt}
\item  $O_L$ can distinguish simple sentences (i.e., there is only one predicate) and complex sentences (i.e., there are two or more predicates). A complex sentence has two kinds: The first kind consists of a simple sentence as a main clause and a few 
subordinate clauses (simple or complex sentences) or sentence segments
with  normalized verbs.
The second kind consists of 
a few independent sentences (simple or complex) connected by conjunction.

\vspace*{-5pt}
\item $O_L$ can segment sentences into a sequence of basic units.
A basic unit could be a phrasal verb, a phrasal noun, or simply a word 
that does not belong to any phrase (if any) contained in the sentence.

\vspace*{-5pt}
\item $O_L$ can assign each basic unit in a sentence with an SR tag and a POS tag.
For a complex sentence of the first kind, $O_L$ can tag the main clause as a simple sentence and each subordinate clause with one SR tag (such as time and cause), and tag each subordinate clause itself
as a sentence. For a complex sentence of the second kind, $O_L$ simply separates the sentence into a collection of individual sentences and tags them accordingly.
Moreover, $O_L$ may be able to produce  other semantic or syntactic tags for each basic unit. 

\vspace*{-5pt}
\item $O_L$ can identify an interrogative pronoun by a POS tag.
An interrogative sentence, however, may or may not include an interrogative pronoun. 
\end{enumerate}

For example, exiting NLP tools for
the English language provide a reasonable approximation to such an oracle. Better approximations are expected when more NLP techniques are developed.
\begin{definition}
Let $k \geq 2$ be a number of tags that $O_L$ can assign to a basic unit.
A $k$-semantic-syntactic unit ($k$-SSU) is a $k$-dimensional vector of tags,
denoted by $(t_1, t_2, \ldots, t_k)$, where $t_1$ is an SR tag, $t_2$ is a POS tag, and $t_i$ ($i>2$) represent other tags of fixed types.
\end{definition}

For example, 
we may add an NE tag to a basic unit to form a 3-SSU; adding one more tag on sentiment  forms a 4-SSU. 
Let $U = (t_1,t_2,\ldots, t_k)$ be an SSU. Denote by $U.i = t_i ~(i \geq 1)$.
The prefix $k$ is omitted when there is no confusion. 

Two consecutive SSUs $A$ and $B$ with $A.1 = B.1$ (i.e., they have the same SR tag)
and $A$ appearing on the left side of $B$ in a sentence may be merged to a new SSU $C$ as follows:
(1) If $A = B$, then set $C \leftarrow A$.
(2) Otherwise, based on the underlying language $L$, either set $C.2 \leftarrow A.2$ (i.e., use the POS tag on the left) or set $C.2 \leftarrow B.2$.
For the rest of the tags in $C$, select a corresponding tag in $A$ or $B$ according to $L$. The following proposition is evident:

\begin{proposition} \label{prop:1}
For any sequence of SSUs, after merging, the new sequence of SSUs does not
have two consecutive SSUs with the same SR tag.
\end{proposition}

To accommodate the situation without proper segmentation of phrasal verbs (see Section \ref{sec:4.7}), 
it is desirable to allow a fixed number of consecutive SSUs to have the same SR tag
in a meta sequence. 
 
\begin{definition} \label{def:2}
A {\it meta sequence} is a sequence  of SSUs such that 
each SR tag appears at most $r$ times,
with interrogative pronouns (if any) left as is without tagging, where
$r \geq 1$ is a positive constant. 
\end{definition}

We assume the availability of \textsl{sentence segmentation} that can segment
a complex sentence to form simple sentences for each clause (main and subordinate),
and we treat such a sentence as 
a set of simple sentences. If a clause itself is a complex sentence, it can be further
segmented as a set of simple sentences. A declarative sentence consists of at least three different SR tags corresponding to
subject, object, and predicate.

Since a complex sentence can be treated as a list of simple sentences, 
MetaQA learns meta sequences 
of declarative sentences and the corresponding interrogative sentences from a training dataset
consisting of such pairs of sentences, where a declarative sentence is a simple
sentence. 

However, there are complex sentence that are not easily segmented into
a set of simple sentences using the existing NLP tools. To represent this type of complex sentences, we may
define a meta sequence as a recursive list of SSUs with a tree structure to represent a sentence using the notion of \textsl{list} in the LISP programming language. This
will be addressed in a separate paper.

\begin{figure}[h]
  \centering
  \includegraphics[width=0.9 \linewidth]{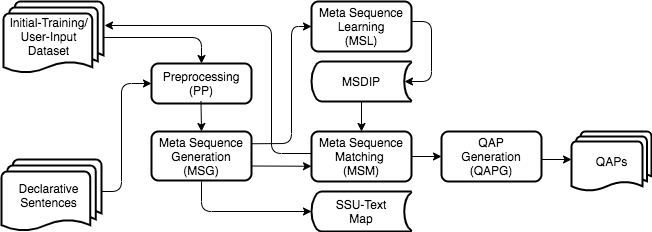}
  \caption{MetaQA architecture and data flow}
  \label{fig:1}
\end{figure}

MetaQA consists of two phases: learning and generation. In the learning phase,
MetaQA learns meta sequence pairs from an initial training dataset to generate an initial MSDIP.
In the generation phase, it takes a declarative sentence as input and 
generates QAPs using MSDIP.
Figure \ref{fig:1} depicts the general architecture and data flow of MetaQA,
which consists of six components: Preprocessing (PP), Meta Sequence Generation (MSG), Meta Sequence Learning (MSL), 
Meta Sequence Matching (MSM), and QAP Generation (QAPG) (see Section \ref{sec:4} for detailed explanations of these components in connection to an implementation of
the English language). 

Both phases use the same PP and MSG components. The PP component is responsible for
tagging basic units in a given sentence (declarative or interrogative) with SR tags, POS tags, and other syntactic and semantic tags, and segmenting complex sentences into a set of simple sentences using oracle $O_L$. The MSG component is responsible for merging SSUs to form a meta sequence. Moreover, for an input sentence in the generation phase, MSG also maps each SSU after merging to the underlying text. 

\subsubsection*{Learning Phase}  

The MSL component removes redundant meta sequences for each pair of
MD and MI generated from MSG and stores the remaining pairs in the MSDIP database.  
Recall that an interrogative pronoun identified by POS tag in an MI is left as is without using its SSU


Note that for any language, $k$ is a constant, 
so are the number of SR tags, the number of POS tags, and
the number of other tags. 
The following proposition is straightforward.

\begin{proposition} \label{prop:2}
(1) For a given language, the length of a meta sequence is bounded above
by a constant, so is the size of MSDIP. 
(2) The length of a meta sequence for a declarative sentence
is at least 3. 
\end{proposition}

\subsubsection*{Generation phase} 
Let $M$ be a meta sequence. Denote by
$M'$ the set of SSUs contained in $M$ and $|M|$ the size of $M'$.
After MetaQA is trained, it generates QAPs from
a given declarative sentences $s$ using the following
{\it QAP-generation algorithm}, where
$X_s$ is
the meta sequence for $s$ generated from MSG. Recall that the text
for each SSU is stored in the SSU-Text Map. 

Step 1. Find a meta sequence MD $X$ from (MD, MI) pairs in MSDIP 
that is the \textsl{best match} of $X_s$. 
This means that the longest common substring of $X$ and $X_s$,
denoted by $Z= \text{LCS}(X,X_s)$, is the longest 
among all MDs in MSDIP.  A substring is a sub-sequence of consecutive SSUs.
If $Z$ contains SSUs for, respectively, a subject, a predicate, and an object, then
we say that it is a \textsl{successful} match.
If furthermore, $Z = X = X_s$, then we say that it is  
a \textsl{perfect} match. If $Z$ is missing a subject SSU, a predicate SSU, or
an object SSU, then it is an \textsl{unsuccessful} matching.
If a match is successful, got Step 2.
If a match is unsuccessful or successful but not perfect, then notify the user that 
MetaQA needs to learn a new pattern and
ask for interrogative sentences for $s$ from the user. After this, go to Step 2.


Step 2.
The goal is to generate all possible interrogative sentences for $s$.  
For each pair $(X,Y) \in \text{MSDIP}$,
generate a meta sequence $Y_s$ from $Y$ with
$$Y'_s = [Y'-(X' \cap Y'-X'_s)] \cup (X'_s - Z').$$
This means that $Y'_s$ is obtained from $Y'$ by removing SSUs that are in both
meta sequences in the matched pair
but not in the input sentence,
and adding SSUs in the input sentence but not in the matched MD.
Since $Z = \text{LCS}(X,X_s)$, the following proposition is straightforward:

\begin{proposition} \label{prop:3}
$X'_s - Z' = X'_s - X'$.
\end{proposition}
Order SSUs in $Y'_s$ appropriately to form $Y_s$, which requires
localization according to the underlying language.
If an SSU in $Y'_s$ has the corresponding text stored in Step 1,
then replace it. If not, then it requires localization to resolve it.
This generate an interrogative sentence $Q_{s}$ for $s$.

Step 3. For each interrogative sentence $Q_{s}$ generated in Step 3,
the SSUs in $A'_s = X' - Y'$ represent a correct answer.
Place SSUs in $A'_s$ in the same order as in $X'_s$ and replace each SSU with the corresponding text in $s$ to obtain an answer $A_s$ for $Q_{s}$.

\section{An Implementation of MetaQA for English} \label{sec:4}

SR, POS, and NE tags are used in this implementation. Existing
NLP tools for generating these tags are for words, not for phrases. 
We could, however, use phrase segmentation to resolve this by appropriate merging operations.
While word segmentation is not needed 
in alphabetic languages such as English, 
phrase segmentation provides a better interpretation of the underlying sentence.
We first assume the existence of an ideal phrase segmentation for English,  
and then discuss how to get around it at the end of this section.

\subsection{Preliminaries}
The following NLP tools are used to generate tags: 
Semantic-Role Labeling (SRL) \cite{shi2019simple} for SR tags,
POS Tagging \cite{toutanova2003feature} for POS tags, and
Named-Entity Recognition (NER) \cite{peters2017semi}
for NE tags.

SR tags are defined in PropBank\footnote{https://verbs.colorado.edu/~mpalmer/projects/ace/EPB-annotation-guidelines.pdf} \cite{bonial2012english,martha2005proposition}, 
which consist of three types: ArgN (arguments of predicates), ArgM (modifiers or adjuncts of the predicates) , and V (predicates).
ArgN consists of six tags: ARG0, ARG1, $\ldots,$ ARG5, and
ArgM consist of
multiple subtypes such as
LOC as location, EXT as extent, DIS as discourse connectives, 
ADV as general purpose, NEG as negation, MOD as modal verb, CAU as cause, TMP as time, PRP as purpose, MNR as manner, GOL as goal, and DIR as direction.

POS tags$\,$\footnote{https://www.ling.upenn.edu/courses/Fall\_2003/ling001/penn\_treebank\_pos.html} are defined in the Penn Treebank tagset \cite{toutanova2003feature,marcus1993building}. For example, NNP is for singular proper noun,
VBZ for third-person-singular-present-tense verb,
DT for determiner, and IN for preposition or subordinating conjunction.

NE tags include PER for persons, ORG for organization, LOC for locations, and numeric expressions for time, date, money, and percentage.

\subsection{PP, MSG, and MSL Localization}

The PP, MSG, and MSL components, on top of what is described in
Section \ref{sec:3}, incur the following localization.
%
%
%
PP first replaces contractions and slang with
words or phrases to help improve tagging accuracy.
For example,  contractions \textsl{'m, 's, 're, 've, n't, e.g., i.e., a.k.a.} are replaced by,
respectively, \textsl{am, is, are, have, not, for example, that is, also known as}.
Slang \textsl{gonna, wanna, gotta, gimme, lemme, ya}  are replaced by,
respectively, \textsl{going to, want to, got to, give me, let me, you}.

PP then segments sentences and
tags words in sentences using SRL, POS Tagging, and NER
for the training dataset and later for input sentences for generating QAPs.
Use SRL to segment a complex sentence into a set of simple sentences and discard all simple sentences without 
a subject or an object. Note that there are complex sentences that are hard to segment using SRL.
Moreover, for each sentence, PP removes
all the words with a CC (coordinating conjunction) as POS tag before its subject, including
\textsl{and, but, for, or, plus, so, therefore}, and \textsl{because}.




MSG then merges the remaining SSUs if two consecutive SSUs are identical.
If they are not identical but have the same SR tag,
then use this SR tag in the merged SSU, and
the POS tag in the first SSU from the right. 
If they contain a noun, use the first SSU from the right with a noun POS tag.
Moreover, the NE tag in the merged SSU is null
if both SSUs contain a null NE tag; otherwise,
use the first non-empty NE tag from the right. 



\subsection{MSM Localization}
The MSM component takes a meta sequence $X_s$ of a sentence $s$ as input 
and executes Step 1 in the
QAP-generation algorithm described in Section \ref{sec:3} using Ukkone's Suffix-Tree algorithm \cite{ukkonen1985algorithms} to compute a longest common substring of two
meta sequences, which runs in linear time. During matching, the POS tags for various types of nouns 
are treated equal; they are NN, NNP, NNS, and NNPS, The POS tags 
for third-person-singular-present verbs are treated equal; they are VBP and VBZ.
To use Ukkone's algorithm,,
we encode a meta sequence as a sequence of symbols using $/$ to separate tags in an SSU.
That is, vector $(t_1,t_2,t_3)$ is now written as $t_1/t_2/t_3$. If $t_2$ is null, then write it as
$t_1//t_3$. If $t_3$ is null, then write it as $t_1/t_2/$. If both are null, then write it
as $t_1//$. SSUs in a sequence are just written as concatenation. For example,
the sentence ``Abraham Lincoln the 16th president of the United States" has
the following SSUs:

\textrm{Abraham (ARG1/NNP/PER) Lincoln (ARG1/NNP/PER) was (V/VBZ/) the (ARG2/DT/)
16th (ARG2/JJ/) president (ARG2/NN/) of (ARG2/IN/) the (ARG2/DT/) United (ARG2/NNP/LOC)
States (ARG2/NNP/LOC)}.

The meta sequence for this sentence is, after merging: 
\textrm{ARG1/NNP/PER V/VBZ/ ARG2/NNP/LOC}.

Let $X$ be an MD in MSDIP such that LCS($X,X_s$) is the longest
among all MDs in MDDIP,
denoted by $Z$. 


\subsection{QAPG Localization}

The QAPG component executes Steps 2--3 in the QAP-generation algorithm described in
Section \ref{sec:3}. Recall that $Z = \text{LCS}(X,X_s)$ is the longest match
among all MDs in MSDIP, and
after 
the set of SSUs $Y'_s$ is generated, localization is needed to form $Y_s$.

\textsl{Case 1:} $Z=X_s$. Then $Y_s = Y$. 

\textsl{Case 2:} $Z$ is a proper substring of $X_s$. Then each SSU in $X'_s - Z$ appears either 
before $Z$ or after $Z$. Form a string $Y_b$ and $Y_a$ of the SSUs that appear,
respectively, before and after $Z$ in the same order as they appear in $X_s$.
Let $Y_s =  [Y-(X'\cap Y'-X'_s)]Y_aY_b$, where $Y-(X'\cap Y'-X'_s)$
means to remove from $Y$ the SSUs in $X'\cap Y'-X'_s$.
 
For each SSU in $Y_s$ if a corresponding text can be found in the SSU-Text Map,
then replace it with the text. An SSU that doesn't have a matched text in the
SSU-Text Map is
due to the helping verbs added in the interrogative sentence that generates $Y$.
%
There are five POS tags for verbs: VBG for gerund or present participle,
VBD past tense, VBN past participle, VBP non-3rd person singular present,
and VBZ 3rd person singular present.
Present participle and past participle have already included helping verbs, and so do the negative forms of
past tense and present tense.
Thus, only positive forms of past tense (VBD) and present tense (VBP, VBZ) 
do not include
helping verbs, which need to be resolved.

\subsubsection*{Rule for resolving helping verbs}

The first V-SSU in $Y$ (i.e., the SSU that contains the SR tag of V) is a helping verb. To determine its form, 
check the POS tag in the subject SSU (usually it is ARG0)
and determine if it is singular or plural. Then check the POS tag in the first V-SSU in $Y$ 
to determine the tense. Replace the second V-SSU with the verb in its original form 
for the V-SSU in the SSU-Text MAP. 

For example, suppose that the following declarative sentence ``John traveled to Boston last week" and its
interrogative sentence about location ``Where did John travel to last week"
are in the training dataset, which generate the following SSUs before merging:

John (ARG0/NNP/PER) traveled (V/VBD/) to (ARG1/IN/) Boston (ARG1/NNP/LOC) last (TMP/NN/) week (TMP/NN/).

Where (Where) did (V/VBD) John (ARG0/NNP/PER) travel (V/VB/) to (ARG1/IN/) 
last (TMP/NN/) week (TMP/NN/)?

Since ``travel to" is a phrasal verb, after merging, we have

John (ARG0/NNP/PER) traveled  to (V/VBD/) Boston (ARG1/NNP/LOC) last week (TMP/NN/).

Where (Where) did (V/VBD) John (ARG0/NNP/PER) travel to (V/VB/) last week (TMP/NN/)?

The following meta-sequence pair $(X,Y)$ is learned for MSDIP:

X  = \textrm{ARG0/NNP/PER V/VBD/ ARG1/NNP/LOC TMP/NN/}

Y = \textrm{Where V/VBD/ ARG0/NNP/PER V/VB/ TMP/NN/}

Suppose that we are given a sentence $s=$ ``Mary flew to London last month." Its meta sequence $X_s$ is exactly the same as $X$, with
ARG0/NNP/PER for ``Mary",
V/VBD/ for ``flew to", ARG1/NNP/LOC for ``London", and TMP/NN/ for ``last month", which
are stored in the SSU-Text Map.
Thus, $Y_s = Y$. 
We can see that 
the SSU of V/VB/ 
in $Y$ is not in the SSU-Text Map.
To resolve the unmatched V/VB/, check the POS tag in the ARG0-SSU, which is NNP, indicating a singular noun.
The POS tag in the first V-SSU is VBD, indicating past tense. Thus, the correct form of the helping verb is ``did". The text for V/VBD is ``flew to" in the SSU-Text Map. The original form of the verb is ``fly".
Thus, the second V-SSU is replaced with ``fly". 
This generates the following interrogative sentence: ``Where did Mary fly to last month?"
The answer SSU is $X'-Y'$, which is ARG1/NNP/LOC, corresponding to ``London".

\subsection{SSU Merging without Segmentation} \label{sec:4.7}

To the best of our knowledge, no tools exist at this point that can segment English sentences to identify phrasal nouns and phrasal verbs. It is worth mentioning that AutoPhrase \cite{shang2018automated} 
can be used for identifying certain phrasal nouns. 
We could deal with 
phrasal verbs using a list of common phrasal verbs or by modifying merging operations.
A phrasal verb consists of a preposition or an adverb, or both. 
There are four POS tags  IN for preposition or subordinating conjunction,
RB for adverb, RBR for comparative adverb, and RBS for superlative adverb.

To see this problem, let us look at the same example aforementioned. After merging,
we have

John (ARG0/NNP/PER) traveled (V/VBD/) to Boston (ARG1/NNP/LOC) last week (TMP/NN/).

Where (Where) did (V/VBD/) John (ARG0/NNP/PER) travel (V/VB/) to (ARG1/IN/) last week (TMP/NN/)?

For the input sentence we have

Mary (ARG0/NNP/PER) flew (V/VBD/) to London (ARG1/NNP/LOC) last moth (TMP/NN/).

The interrogative sentence is ``Where did Mary fly ARG1/IN/ last week?" after replacing SSUs with text in
the SSU-Text Map, with ARG1/IN/ unmatched with text.
We can resolve this by modifying the merging operation as follows:
When an SSU with a POS tag for preposition or adverb appears appears before or after a V-SSU, leave it as is without merging it with its neighboring SSUs of the same SR tag, unless the POS tags 
in them are also for prepositions or adverbs. The rest of the merging operations are the same. Then we have, after merging,

John (ARG0/NNP/PER) traveled (V/VBD/) to (ARG1/IN/) Boston (ARG1/NNP/LOC) last week (TMP/NN/).

Where (Where) did (V/VBD/) John (ARG0/NNP/PER) travel (V/VB/) to (ARG1/IN/) last week (TMP/NN/)?

Now the input sentence becomes, after SSU merging,

Mary (ARG0/NNP/PWR) flew (V/VBD) to (ARG1/IN/) London (ARG1/NNP/LOC) last moth (TMP/NN/).

All the SSUs in the meta sequence  ``Where V/VBD/ ARG0/NNP/PER V/VB/ ARG1/IN/ TMP/NN/" have corresponding text 
 in the SSU-Text Map after resolving for helping verbs.  The answer
SSU is in $X'-Y'=$ ARG1/NNP/LOC, which is ``London".

\section{Evaluations} \label{sec:5}

To evaluate MetaQA, we need to have appropriate evaluation measures,
training data, and evaluation data.
BLUE \cite{10.3115/1073083.1073135}, ROUGE \cite{lin-2004-rouge}, and Meteor \cite{10.1007/s10590-009-9059-4} are standard evaluation metrics for measuring automatic summarization and machine translation, which are good for computing text similarity
and have also been used to evaluate QG. Another commonly-used measure is 
human judgments.
BLEU and ROUGE-N count the number of overlapping units between the candidate text and the reference text by using N-grams.
ROUGE-L measures the cognateness between the candidate text and the reference text by using Longest common sub-sequence.
Meteor compares the candidate text with the reference text in terms of exact, stem, synonym, and paraphrase matches between words and phrases.
These metrics, however, do not evaluate grammatical correctness. Thus, human judgment 
is the only liable measure for grammatical correctness.

SQuAD \cite{Rajpurkar_2016} is a  dataset that has been
used for training and evaluating generative methods for QG.
However, not all QAPs in SQuAD are well-formed or with correct answers. 
There are also about 20\% of questions in the dataset that require paragraph-level information.
Thus, SQuAD is unsuitable for evaluating QAPs for our purpose.
Instead, we constructed an initial training dataset by writing a number of declarative sentences
and the corresponding interrogative sentences to cover the major
tense, participles, voice, modal verbs, and some common phrasal verbs such as
``be going to" and ``be about to" for the following six interrogative pronouns: 
\textsl{Where, Who, What, When, Why, How many}. A total of 112 meta-sequence pairs (MD, MI) were learned as  the initial MSDIP.

To evaluate MetaQA, we extracted declarative sentences from the official SAT practice reading tests$\,$\footnote{https://collegereadiness.collegeboard.org/sat/practice/full-length-practice-tests}, for the reason that
SAT practice reading tests provide a large number of different patterns of declarative sentences. 
 There are a total of eight SAT practice reading tests, each consisting of five articles and
 each article consisting of around 25 sentences, for a total of 40 articles and 1,136 sentences. 
After removing easy-to-identify interrogative sentences and imperative sentences,
we harvested a total of 1,025 sentences (which may still contain imperative
sentences).
Using the initial MSDIP, MetaQA generated a total of 796 QAPs.

Three native Chinese speakers evaluated the QAPs on a shared
Google doc file based on the following criteria:
For questions: Check both syntax and semantics: (1) correct; (2) acceptable (e.g., a minor would make it correct); (3) not acceptable.
 For answers: (1) matched---the answer matches well with the question; (2) acceptable; (3) not acceptable.
 The final results were agreed by the three judges. Presented below 
 are questions generated with detailed breakdowns in each
category, where ``all correct" means both syntactically and semantically correct and conforming to native-speaker norms,
``not acceptable" means either syntactically or semantically unacceptable, and
``How" means ``How many": 

\medskip
\begin{center}
\begin{tabular}{l|c|c|c|c|c|c|c}
\hline
& \textbf{Where} & \textbf{Who} & \textbf{What} & \textbf{When} & \textbf{Why} & \textbf{How} & \textbf{Total} \\
\hline
MSDIP pairs & 18 & 45 & 23 & 22 & 6 & 8 & 122 \\
\hline
QAPs generated & 26 & 216 & 466 & 51 & 15 & 22 & 796 \\
\hline
All correct & 21 & 208 & 458 & 51 & 15 & 20 & 773 \\
\hline
Syntactically acceptable & 4 & 4 & 3 & 0 & 0 & 2 & 13 \\
\hline
Semantically acceptable & 1 & 2 & 5 & 0 & 0 & 0 & 8	 \\
\hline
Not acceptable & 0 & 2 & 0 & 0 & 0 & 0 & 2 \\
\hline
\end{tabular}
\end{center}
\smallskip
The percentage of generated questions that are both syntactically and semantically correct
is 97\%. 
We noticed that there is a strong correlation between
the correctness of the questions and their answers. In particular, 
when a generated question is all correct, its answer is also all correct.
When a question is acceptable, its answer may be all correct or acceptable.
Only when a question is unacceptable, its answer is also unacceptable.
 

The 13 incorrect but syntactically acceptable questions are mostly due to some minor issues
in segmenting a complex sentence into simple sentences, where a better
handling of sentence segmentation is expected to correct these issues. Two questions whose
interrogative pronoun should be ``how much" are mistakenly using ``how many".
Further refinement of POS tagging that distinguish uncountable nouns from countable nouns would solve this problem. The eight semantically acceptable questions are all due to 
NE tags that cannot distinguish between persons, location, and things. Further refinement of NE tagging will solve this problem. The two unacceptable questions are due to
serious errors induced when segmenting complex sentences. This suggests that we should
look into using a recursive list to represent complex sentences.

There were 589 sentences for which no matched meta sequences are found from 
the initial MSDIP. By learning new meta sequences from user inputs,
535 of these sentences found perfect matching, which generate
QAPs that are both syntactically and semantically correct. For the remaining 84 sentences,
some of then are imperative sentences without a clear structure of subject-predicate-object,
and some are hard to segment into a set of simple sentences due to inaccurate SR tagging
and so no
appropriate (MD, MI) pairs were learned. This suggests that we should
look into better sentence segmentation methods or meta trees as recursive lists of meta sequences to represent complex sentences
as a whole, which is left for future work.

We evaluated the running time to generate QAPs over 100 sentences on a desktop computer with an Intel Core I5 2.6 Ghz CPU and 16 GB RAM. The average running time is 0.55 seconds for each input sentence, which is deemed satisfactory for online applications. For a given article, assuming that it would take the reader several minutes to read. By then all the QAPs for MCQs would have been generated.

\section{Conclusions and Final Remarks} \label{sec:6}

Meta sequence learning is a novel approach for generating adequate QAPs,
which  
achieves satisfactory results for the English language using existing NLP tools on
SR, POS, and NE tagging. 
Further improvement of named-entity recognition may be able to eliminate a small number of semantic errors we encountered in our evaluations. 
When almost all possible patterns of declarative sentences and the corresponding interrogative sentences are learned (there are only finitely many of them to be learned), MetaQA is expected to perform well on generating adequate QAPs from declarative sentences that can be segmented appropriately into simple sentences.

However, not all complex sentences can be segmented using
existing tools. In particular, about 7.4\% of the declarative sentences in the 
official SAT practice reading tests are in this category. This calls for, as mentioned
near the end of Section \ref{sec:5}, a better NLP method to help dissect complex sentences. Another approach to resolving this is to use a tree structure of meta sequences. For example, we may be able to represent a complex sentence as a recursive list of meta sequences. 

Applying MetaQA to a logographic languages would require robust and accurate segmentation at all levels of words, phrases, and sentences, semantic labeling, POS tagging, and named-entity recognition for the underlying languages.
It would also require appropriate  localization for
merging SSUs. 
It is interesting to explore how well MetaQA performs on a language other than English.

Generating QAPs on derived points remains a challenge. It calls for a major breakthrough in machine inference.
It would be interesting to investigate how meta sequence learning may help generate such QAPs. Incorporating neural-network technologies and meta sequence learning could be
a direction worth exploring. Adding more semantic tags and syntactic tags may also help, such as sentiment tags and logic tags to carry out certain forms of reasoning. 
\bibliographystyle{coling}
\bibliography{QG-reference}

\end{document}